\documentclass[a4paper,twoside]{article}

\usepackage{epsfig}
\usepackage{subcaption}
\usepackage{calc}
\usepackage{amssymb}
\usepackage{amstext}
\usepackage{amsmath}
\usepackage{amsthm}
\usepackage{multicol}
\usepackage{pslatex}
\usepackage{apalike}
\usepackage{algorithm2e}
\usepackage[bottom]{footmisc}
\usepackage{SCITEPRESS}     

\begin{document}

\title{Few-Shot Histopathology Image Classification: Evaluating State-of-the-Art Methods and Unveiling Performance Insights}

\author{\authorname{Ardhendu Sekhar\sup{1}, Ravi Kant Gupta\sup{2} and Amit Sethi\sup{3}}
\affiliation{\sup{1}Department of Electrical Engineering, Indian Institute of Technology Bombay, Mumbai, India}
\email{\{214070020, 184070025, asethi\}@iitb.ac.in}
}


\abstract{This paper presents a study on few-shot classification in the context of histopathology images. While few-shot learning has been studied for natural image classification, its application to histopathology is relatively unexplored. Given the scarcity of labeled data in medical imaging and the inherent challenges posed by diverse tissue types and data preparation techniques, this research evaluates the performance of state-of-the-art few-shot learning methods for various scenarios on histology data. We have considered four histopathology datasets for few-shot histopathology image classification and have evaluated 5-way 1-shot, 5-way 5-shot and 5-way 10-shot scenarios with a set of state-of-the-art classification techniques. The best methods have surpassed an accuracy of 70\%, 80\% and 85\% in the cases of 5-way 1-shot, 5-way 5-shot and 5-way 10-shot cases, respectively. We found that for histology images popular meta-learning approaches is at par with standard fine-tuning and regularization methods. Our experiments underscore the challenges of working with images from different domains and underscore the significance of unbiased and focused evaluations in advancing computer vision techniques for specialized domains, such as histology images.}

\onecolumn \maketitle \normalsize \setcounter{footnote}{0} \vfill

\section{\uppercase{Introduction}}
\label{sec:introduction}

 Traditional deep learning models often require large amounts of labeled data for training. These models learn representations and patterns from a substantial dataset to generalize well to unseen examples. The learning process involves adjusting numerous parameters through backpropagation to minimize the difference between predicted and actual outputs. Transfer learning is commonly used in traditional deep learning, where models pre-trained are fine-tuned for specific tasks. The knowledge gained from the pre-training on large dataset helps in solving related problems with smaller dataset. In medical imaging, traditional deep learning methods have been successful for tasks such as image classification and segmentation, but they often require extensive labeled dataset. But traditional models might struggle when faced with new tasks or limited data, requiring substantial retraining or fine-tuning. This is where few-shot learning can be useful. Few-shot learning, as the name implies, is designed to make accurate predictions when only a small number of examples per class are available for training. This can be achieved through various techniques such as meta-learning, where the model is trained on a variety of tasks and adapts quickly to new tasks with minimal examples by leveraging knowledge gained from previous tasks. One would think that meta learning would be particularly valuable in medical imaging due to the scarcity and cost of labeling medical data. However, such techniques have only been evaluated on natural images. If successful on medical images, this would enable demonstrate learning even when only a handful of annotated medical images are available.
 
In this work, we have evaluated certain state-of-the-art few-shot classification techniques on histopathology medical datasets. A dataset prepared by \cite{komura2021histology} and FHIST \cite{shakeri2022fhist} were considered for the experiments. FHIST dataset comprises of many histology datasets. These are : CRC-TP~\cite{9204851}, NCT-CRC-HE-100K \cite{kather2019predicting}, LC25000 \cite{borkowski2019lung} and BreakHis \cite{7312934}. For our experiments, we have considered CRC-TP, NCT and LC25000 datsets. CRC-TP is a colon cancer dataset with six classes. NCT is also a colon cancer dataset with nine classes. LC25000 consists of both lung and colon cancer images with five classes. We have also used a histology dataset proposed by \cite{komura2021histology}. It has around 1.6 million cancerous image patches of 32 different organs in the body. The classes in this dataset are defined according to the different organ sites. Dataset proposed by \cite{komura2021histology} is used to train the few-shot classification models. The trained model is then evaluated on various FHSIT dataset.

\begin{figure*}[htbp]
\centering
\includegraphics[height=5cm,width=10cm]{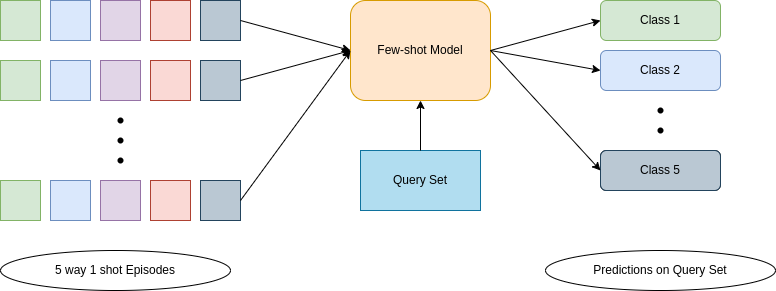}
\caption{The diagram illustrates a few-shot learning model, showcasing its ability to effectively generalize and recognize classes for unlabeled query set with only a limited number of support examples.The 5 different colors in support set represent 5 different classes(ways) having 1 sample(shot) each. }
\label{fig1}
\end{figure*}

In our comprehensive exploration of few-shot classification techniques, we have meticulously incorporated various state-of-the-art methodologies to ensure a thorough evaluation. The methods employed in our experiments includes Prototypical Networks \cite{snell2017prototypical}, Model-Agnostic Meta-Learning (MAML) \cite{finn2017modelagnostic}, SimpleShot \cite{wang2019simpleshot}, LaplacianShot \cite{ziko2020laplacian},  and DeepEMD \cite{zhang2020deepemd}. Prototypical Networks, a benchmark technique for few-shot learning, leverage prototypes as representative embeddings for each class. By minimizing the distance between query examples and class prototypes, this model excels in adapting quickly to new classes with limited labeled data. MAML adopts a meta-learning paradigm, training a model to quickly adapt to new tasks with minimal examples. This approach proves invaluable in scenarios where prompt adaptation to novel classes is paramount. As its name implies, SimpleShot emphasizes simplicity in few-shot learning. This approach often relies on straightforward yet effective techniques, showcasing the power of simplicity in addressing complex classification tasks with limited data. LaplacianShot introduces Laplacian regularization to enhance few-shot learning performance. By incorporating this regularization technique, the model aims to improve generalization and robustness across diverse classes. DeepEMD learns the image representations by calculating the discrepancy between the joint characteristics of embedded features and product of the marginals.

\section{\uppercase{Related Work}}

In recent years, there has been a notable surge in research endeavors for addressing the challenges of few-shot learning within specific domains of medical imaging. This section outlines some contributions in the field, shedding light on how researchers have applied various few-shot learning techniques to tackle different problems in medical diagnostics. One such study \cite{mahajan2020metadermdiagnosis}, focuses on skin-disease identification. Here, the researchers applied two prominent few-shot learning methods, Reptile \cite{nichol2018reptile}  and Prototypical Networks \cite{snell2017prototypical}, showcasing their efficacy in the task of distinguishing various skin diseases. The task of transferring knowledge across different tissue types was addressed in \cite{medela2019fewshot}, where a deep Siamese neural network was trained to transfer knowledge from a dataset containing colon tissue to another encompassing colon, lung, and breast tissue. Meanwhile, \cite{teng2021fewshot} proposed a few-shot learning algorithm based on Prototypical Networks, specifically tailored for classifying lymphatic metastasis of lung carcinoma from Whole Slide Images (WSIs). In another work in \cite{chen2021momentum}, a two-stage framework was adopted for the crucial task of COVID-19 diagnosis from chest CT images. The initial stage involved capturing expressive feature representations using an encoder trained on publicly available lung datasets through contrastive learning. Subsequently, in the second stage, the pre-trained encoder was employed within a few-shot learning paradigm, leveraging the Prototypical Networks method. In \cite{yang2022towards}, the authors have incorporated contrastive learning (CL) with latent augmentation (LA) to build a few-shot classification model. Here, contrastive learning learns important features without needing labels, while Latent Augmentation moves semantic variations from one dataset to another without supervision. The model was trained on publicly available colon cancer datasets and evaluated on PAIP \cite{kim2021paip} liver cancer Whole Slide Images. Shakeri et al. \cite{shakeri2022fhist} introduced the FHIST data coupled with assessments of certain few-shot classification methodologies serves as a catalyst for additional exploration in this domain. 

The exploration of few-shot learning is not confined to image classification alone. In \cite{ouyang2020selfsupervision}, a few-shot semantic segmentation framework was introduced, aiming to alleviate the dependency on labeled data during the training phase. Similarly, \cite{yu2021locationsensitive} ventured into the domain of medical image segmentation, employing a prototype-based method known as the location-sensitive local prototype network, which strategically incorporates spatial priors to enhance segmentation accuracy. 

\section{\uppercase{Methodology}}

Few shot classification deals with the cases of limited data. For example, a medical dataset has five rare classes and in each class there are five images. It is very difficult to train and test DL models with such small number of images. Training the model with such small dataset my lead to over-fitting. Since the classes are rare, there is a high probability that the pre-trained models might not have seen those images while being trained. Instead of traditional transfer-learning, the few-shot models follow episodic-training paradigm.

In few-shot training, the train and test set are disjoint. The training set is denoted as $D_{Train}$. It consists of $(X_{Train},Y_{Train})_{i=1}^{M}$. $X_{Train}$ are the images. $Y_{Train}$ are the labels. M is the number of classes. In episodic training, the huge labelled train dataset is broken into many episodes. In each episode, there is a support set(S) and a query set(Q). Each episode is defined by K-way N-shot Q-query. It means, for each episode, K classes are randomly selected from M classes. From each K class, N images and Q images are selected. N$\times$K images form the support set and Q$\times$K images form the query set. The few shot model is trained in such a way that it must learn from the support set to predict the labels of the query set. Once the model is trained, it is tested on the test set.

\begin{equation}
D_{Train}=(X_{Train},Y_{Train})
\label{eqn1}
\end{equation}
\begin{equation}
S := {(X_{i},Y_{i})_{i=1}^{K \times N}}
\label{eqn2} 
\end{equation}
\begin{equation}
Q := {(X_{i},Y_{i})_{i=1}^{K \times Q}}
\label{eqn3} 
\end{equation}
\begin{equation}
D_{Test}=(X_{Test},Y_{Test})
\label{eqn4}
\end{equation}
\begin{equation}
S := {(X_{i},Y_{i})_{i=1}^{K \times N}}
\label{eqn} 
\end{equation}
\begin{equation}
Q := {(X_{i})}
\label{eqn} 
\end{equation}
\begin{equation}
Y_{Train} \cap Y_{Test} = \Phi
\label{eqn5}
\end{equation}

\subsection{Prototypical Networks}

The central idea of Prototypical Networks \cite{snell2017prototypical} is rooted in the observation that data points exhibiting proximity to a singular prototype representation for each class contribute to a meaningful embedding. To operationalize this concept, a non-linear mapping transforms the input data into a specialized embedded space. Within this embedded space, the prototype representation for each class was derived by calculating the mean of its respective support set. The process is initiated by the formation of a non-linear mapping, effectively transforming the original input data into a specialized embedded space. This mapping is crucial for capturing intricate relationships and patterns within the data. Within the embedded space, a class prototype was defined for each category. The prototype representation for a specific class was computed by determining the mean of its corresponding support set. This means that the prototype becomes a central point representing the class in the embedded space. To perform classification on a given query, the approach relies on identifying the nearest distance between the embedded query and the prototype representation of each class. The class associated with the closest prototype is then assigned to the query, effectively determining its classification.

The small support set consists of K labeled examples which is represented by 
S = [($X_1$,$Y_1$),...,($X_K$,$Y_K$)]. $X_i$ is a D-dimensional feature vector for each image and $Y_i$  represents the corresponding label of $X_i$. There are N classes with in a support set. The number of examples with in a class N is represented by $S_N$.

With the help of an embedding function, i.e. a CNN,$f_{\Phi}$, prototypical networks estimate an M dimensional representation $C_N$ of each class. $C_N$ is the mean vector of the support points of each class. $\Phi$ are the learnable parameters.

\begin{equation}
f_\Phi:R^D -> R^M 
\label{eqn6}
\end{equation}
\begin{equation}
C_N \in R^M
\label{eqn7}
\end{equation}
\begin{equation}
C_N=\frac{1}{|S_N|}\sum_{(X_i,Y_i)\in S}f_{\Phi}(X_i)
\label{eqn8}
\end{equation}

Prototypical networks generate a probability distribution across classes for a given query example X. This distribution is computed using a softmax function applied to the distances between the query example and other prototypes in the embedding space.
\begin{equation}
P_{\Phi}(Y=N|X)=\frac{\exp(-d(f_{\Phi}(X),C_N))}{{\sum_{N^{'}}}\exp(-d(f_{\Phi}(X),C_{N^{'}}))}
\label{eqn9}
\end{equation}

The model learns by minimizing the negative of the log probability of $J(\Phi)$ of the true class N through Adam solver. 

\begin{equation}
J(\Phi)=-\log(P_{\Phi}(Y=N|X))
\label{eqn10}
\end{equation}

\subsection{Model Agnostic Meta Learning(MAML)}

MAML \cite{finn2017modelagnostic} introduces a meta-learning objective that involves training a model on a distribution of tasks. The goal is to learn a good initialization of the model's parameters that facilitates quick adaptation to new, unseen tasks. The focus of MAML is on few-shot learning scenarios, where the model is required to generalize well from a small number of examples per class. This is particularly important in real-world applications where collecting extensive labeled dataset can be impractical. During the meta-training phase, the model is exposed to a variety of tasks. Each task consists of a support set (small labeled dataset) and a query set (unlabeled examples). The model is trained to adapt quickly to new tasks by updating its parameters through gradient descent. MAML introduces a two-step gradient descent update during meta-training. The first update involves computing the gradients on the support set and using them to update the model's parameters. The second update fine-tunes the model on the query set to improve its performance on the specific task. The model's initialization is learned in a way that allows it to adapt quickly to new tasks. The meta-training process encourages the model to learn parameter initialization that generalize well across different tasks. During meta-testing, the model is fine-tuned quickly on new tasks with limited examples. This is achieved by applying a small number of gradient descent steps using the updated initialization. The goal is to improve the model's performance on the specific task.

The neural network is parameterized by a set of weights $\theta$. The weights are to be updated in such a way that they can be rapidly adapted towards different solutions. For instance, if there are 3 episodes $E_1$, $E_2$, $E_3$ then through a few gradient steps the weights should be able to move from a specific point $\theta$ to another configuration of weights $\theta_1^*$, $\theta_2^*$ and $\theta_3^*$. It should be done in such a way that the configuration of weights are now well adapted on the task to be solved. This can be achieved by optimizing three different losses on three different support sets at the same time. In this case, 2 losses will be optimized. The first loss is across all the support to find $\theta$. Another query set specific loss will be optimized in $\theta_1^*$, $\theta_2^*$ and $\theta_3^*$ directions.

\begin{equation}
S={[(X_1,Y_1),(X_2,Y_2)...(X_5,Y_5)]}
\label{eqn11}
\end{equation}
\begin{equation}
Q={[(X_1,Y_1]}
\label{eqn12}
\end{equation}

The two above equations represent a particular episode. Suppose there are 3 different episode. In each of of the support sets of these episodes, there are five different images and five labels. And in each query set, there is one image and one label. The objective is finding out the one particular class to which the query image belongs.It is a standard few shot setting. The below equations will depict how gradient descent happens in one episodic training.

\begin{equation}
E_1 : f_\theta(S_1)= V_1
\label{eqn13}
\end{equation}
\begin{equation}
\theta_1^{*}=\theta-\alpha\Delta_\theta\mathcal{L}(S_1,V_1)
\label{eqn14}
\end{equation}
\begin{equation}
f_{{\theta_1^{*}}}(Q_1)= Z_1
\label{eqn15}
\end{equation}
\begin{equation}
\theta=\theta-\beta\Delta_\theta\mathcal{L}(Q_1,Z_1)
\label{eqn16}
\end{equation}

Considering episode 1, the model is parameterized by $f_\theta$. The $S_1$ is passed through $f_\theta$, $V_1$ is obtained as output.The $\theta$ is fine tuned towards a specific set of parameters $\theta_1$. This $\theta_1$ is used to do another forward pass on the query set. This is the prediction on the query image. The loss is calculated from the query set which is used to do a backward pass on the $\theta$. The task is repeated for the rest other episodes $E_2$ and $E_3$. Then $\theta$ is updated by taking the gradient of accumulated losses of all the query sets in all the tasks.

\subsection{SimpleShot}
SimpleShot \cite{wang2019simpleshot} is a simple non-episodic few-shot learning method. It uses the concept of transfer-learning and nearest-neighbour rule. A large scale deep neural network is trained on training classes. Then nearest-neighbour rule is carried out on the images of the test episodes by using the trained deep neural network as a feature encoder.

Training set is defined as $D_{Train}$.
\begin{equation}
D_{Train}=[(X_1,Y_1),(X_2,Y_2),...,(X_N,X_N)]
\label{eqn17}
\end{equation}

X are the images. Y are the corresponding labels. The train set is trained on a CNN with cross entropy loss.

\begin{equation}
argmin_\theta\sum_{(X,Y)\in D_Train}{l(W^T f_\theta(X),Y)}
\label{eqn18}
\end{equation}

$f_\theta$ represents the CNN. W represents the weights of the classification layer. l represents the cross entropy loss function. 

Classification on the test set is done using the nearest neighbour rule. The features of an image are obtained by passing it through the CNN. The features are denoted by Z. The nearest neighbour classification is carried out using the Euclidean distance.

\begin{equation}
Z=f_\theta(X)
\label{eqn19}
\end{equation}

In one shot setting, support set S of test set has 1 labelled example from each of the N classes.
\begin{equation}
S=((\hat{X}_1,1),...,(\hat{X}_N,N))
\label{eqn20}
\end{equation}

Using the Euclidean distance measure, the nearest neighbour rule classifies the query image $\hat{X}$ to the most similar support image. 

\begin{equation}
Y(\hat{X})=argmin_{N\in(1,2,..,N)}d(\hat{Z},\hat{Z}_N)
\label{eqn21}
\end{equation}

$\hat{Z}$ and $\hat{Z}_N$ are the CNN features of the query and support images respectively. In multi shot setting, $\hat{Z}_N$ is the average of feature vector of each class in the support set.

\subsection{LalplacianShot}

LaplacianShot \cite{ziko2020laplacian} introduces a new approach for few-shot tasks - a transductive Laplacian regularized interference. It minimises a quadratic binary assignment function comprising of two essential terms. The first one is a unary term that allocates query samples to their closest class prototype.  The second term is a pairwise-Laplacian term that promotes consistent label assignments among nearby query samples.

Similar to SimpleShot, train set is defined as $D_{Train}$. The train set is trained on a CNN with a simple cross-entropy loss. It does not involve any episodic or meta learning strategy. 

\begin{equation}
argmin_\theta\sum_{(X,Y)\in D_Train}{l(W^T f_\theta(X),Y)}
\label{eqn22}
\end{equation}

The regularization equations involved at the time of few-shot test inference are described below.

\begin{equation}
E(Y)=N(Y)+\lambda \frac{1}{2} L(Y)
\label{eqn23}
\end{equation}
\begin{equation}
N(Y)=\sum_{q=1}^N \sum_{c=1}^C y_{q, c} d(z_q-m_c)
\label{eqn24}
\end{equation}
\begin{equation}
L(Y)=\frac{1}{2} \sum_{q, p} w(z_q, z_p)||y_q-y_p||_2^2
\label{eqn25}
\end{equation}
In this objective, the first term N(Y) is minimized by assigning each query point top the class of the nearest prototype $m_c$, from the support set, using a distance metric such as Euclidean distance. The second term, L(Y), represents the Laplacian reguralizer  and is expressed as tr($Y^{T}$LY). L is a laplacian matrix that corresponds to the affinity matrix W =  $w(z_q, z_p)$. It measures the similarity between the feature vectors  $z_q$ and $z_p$ with the help of a kernel function. $z_q$ and $z_p$ are feature vectors of the query images $x_p$ and $x_q$.

\subsection{DeepEMD}

In DeepEMD \cite{zhang2020deepemd}, Earth Mover's Distance is used as a metric to calculate the structural distance dense image representations, determining image similarity. The Earth Mover's Distance produces optimal matching flows between structural elements, minimizing the matching cost. The minimized cost is then utilized to signify the image distance for classification. To derive crucial weights for elements in the EMD, a cross-reference mechanism is introduced which mitigates the impact of clustered backgrounds and intra-class variations. For K-shot classification, a structured fully connected layer capable of directly classifying dense image representations using EMD is used.

The Earth Mover's Distance serves as a distance metric between the two sets of weighted objects or distributions, leveraging the fundamental concept of distance between individual objects. It adopts the structure of the Transportation Problem in Linear Programming. A set of supplier S = ({$S_i$$|$i=1,2....m}) must transport goods to a set of demanders D = ({$d_j$$|$j=1,2....k}). $S_{i}$ represents the supply unit of supplier and $d_{j}$ indicates the demand of demander j. The expense fro transporting one unit from supplier i to demander j is represented by $c_{ij}$. And the quantity of units transported is represented by $x_{ij}$. The objective of the transportation problem is to find the least-expensive flow of goods $\tilde{x}$ = ($\tilde{x}_{ij}$$|$i=-1,2,...,m,j=1,2....,k) from supplier i to demander j:

\begin{equation}
\underset{x_{i j}}{\operatorname{minimize}} \quad \sum_{i=1}^m \sum_{j=1}^k c_{i j} x_{i j}
\label{eqn26}
\end{equation}

In few shot classification, to calculate the similarity between the support and query images , the images are passed through a fully convolutional network to generate image features S and Q.

\begin{equation}
\mathbf{S} \in \mathbb{R}^{H \times W \times C}
\label{eqn27}
\end{equation}
\begin{equation}
\mathbf{Q} \in \mathbb{R}^{H \times W \times C}
\label{eqn28}
\end{equation}

Each image feature is a collection of vectors i.e. S = [$s_{1}$,$s_{2}$,....,$s_{HW}$] and  Q = [$q_{1}$,$q_{2}$,....,$q_{HW}$]. The matching cost between the 2 set of vectors is denoted as the similarity of images. The cost between the 2 embeddings $s_{i}$ and $q_{j}$ is given by:  

\begin{equation}
c_{i j}=1-\frac{\mathbf{s}_i^T \mathbf{q}_j}{\left\|\mathbf{s}_i\right\|\left\|\mathbf{q}_j\right\|},
\label{eqn29}
\end{equation}

In a support set, when the shot is more than 1 then the learnable embedding becomes a group of image features for a class rather than one vector. Then the mean of it is taken to give a single image feature. It is similar to calculating the prototype of a class in a support set. The fully connected network is a feature extractor and SGD optimizer is used to update the weights by sampling few-shot episodes from the dataset. 

\begin{figure*}[htbp]
\centering
\includegraphics[height=10cm,width=17cm]{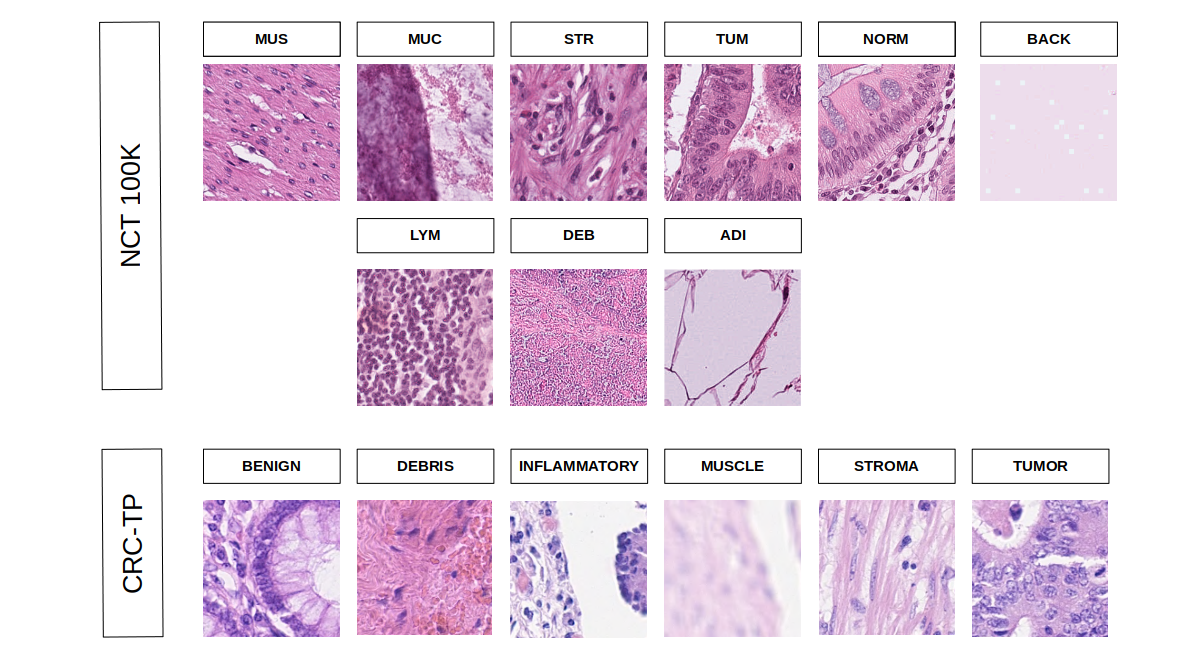}
\caption{Snapshot of sample images of each class from NCT (top 2 rows) and CRC-TP (bottom row) of FHIST dataset.}
\label{fig2}
\end{figure*}

\begin{figure*}[htbp]
\centering
\includegraphics[height=10cm,width=18cm]{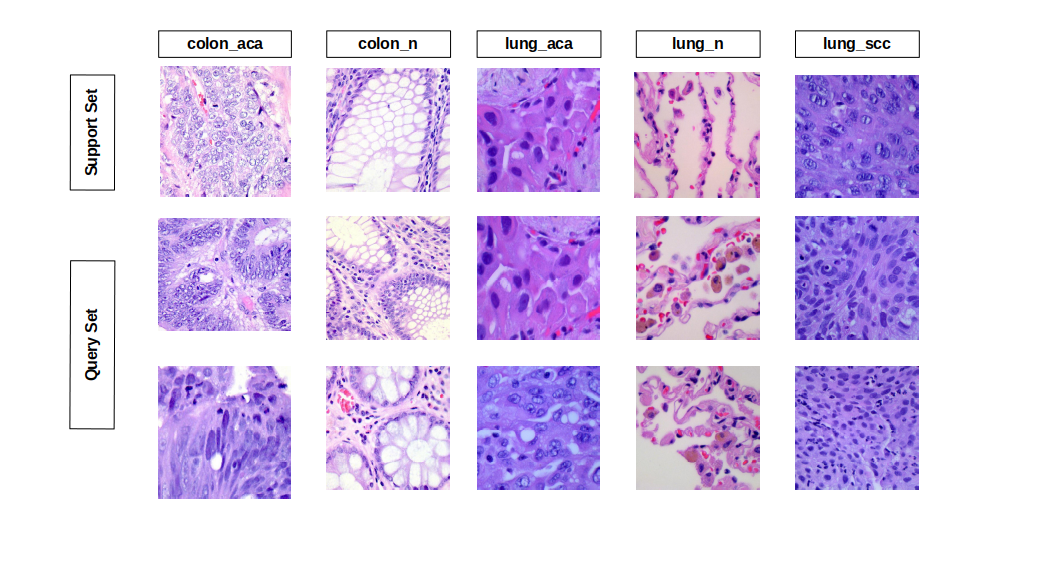}
\caption{An example of 5-way 1-shot 2-query episode. The first row represents the support set from 5 different classes of LC25000 dataset. The last two rows represent the query set.}
\label{fig3}
\end{figure*}

\section{\uppercase{Experiments and Results}}

\subsection{Dataset}

\cite{komura2021histology} have created a dataset of histology images from uniform tumor regions in The Cancer Genome Atlas Program whole slide images\cite{tcga}. TCGA consist of tissue slides from 32 cancer types at different sites of a human body. From this dataset, \cite{komura2021histology} created a set of 1,608,060 image patches with six different magnification levels i.e. 0.5$\mu$/pixel, 0.6$\mu$/pixel, 0.7$\mu$/pixel, 0.8$\mu$/pixel, 0.9$\mu$/pixel, 1.0$\mu$/pixel. As these images are obtained from 32 different types of cancer, they are categorized into 32 classes. NCT \cite{kather2019predicting} dataset comprises of 100,000 image patches of human colorectal cancer extratcted from Hematoxylin and Eosin stained histological images and normal tissue. The resolution of images are 224$\times$224. These images are categorized into seven classes which are Adipose(ADI), background(BACK), debris(DEB), lymphocytes(LYM), mucus(MUC), smooth muscle(MUS), normal colon mucosa (NORM), cancer-associated stroma (STR) and colorectal adenocarcinoma epithelium (TUM).  LC25000 \cite{borkowski2019lung} or Lung and Colon Histopathological dataset contains 25000 image patches, As the name suggests, it comprises of images from lung and colon cancer. The resolution of the images is 768$\times$768. they are categorized into five classes. Three classes belong to lung cancer and the remaining two classes belong to colon cancer. The classes are benign colon tissues, colon adenocarcinoma, lung squamous cell  carcinoma and benign lung tissues. CRC-TP \cite{9204851} is also a colon cancer dataset. It consists of 280,000 image patches categorized into six classes. They are tumor, stroma, complex stroma muscle, debris, inflammatory and benign. The images dimensions are 150$\times$150. The dataset prepared by Komura et al. is used as training dataset. As the dataset is huge, a formidable few-shot backbone network can be expected after training on it. Rest other dataset are used as testing set. In few-shot learning, the test and training set are disjoint.

\subsection{Results and Discussion}

All the experiments are conducted on an NVIDIA A100 in pytorch. Two types of training regimes are followed: Episodic training and standard training. Standard training trains across an entire dataset with abundant examples per class. In contrast, episodic training, designed for few-shot learning, adapts a model to quickly generalize from small episodes containing very few examples of new classes. Models such as Protonet \cite{snell2017prototypical}, MAML \cite{finn2017modelagnostic}, DeepEMD \cite{zhang2020deepemd} follow episodic training. SimpleShot \cite{wang2019simpleshot} and LaplacianShot \cite{ziko2020laplacian} follow the standard training procedure.

\begin{table*}
\caption{Table 1: Accuracy(\%) on three different datasets; CRC-TP \cite{9204851}, NCT \cite{kather2019predicting} and LC25000 \cite{borkowski2019lung}. The few-shot models are trained on the dataset proposed by \cite{komura2021histology}}
\begin{center}
\begin{tabular}{lcccc}
\hline  & \multicolumn{3}{c}{ CRC-TP } \\
Method &  Training Method &  5-way 1-shot & 5-way 5-shot & 5-way 10-shot \\
\hline MAML \cite{finn2017modelagnostic} & Episodic & $38.5$ & $58.7$ & $63.0$ \\
ProtoNet \cite{snell2017prototypical} & Episodic & $43.8$ & $63.6$ & $68.3$ \\
DeepEMD \cite{zhang2020deepemd} & Episodic & $47.3$ & $64.6$ & $68.6$ \\
SimpleShot \cite{wang2019simpleshot} & Standard & $47.9$ & $66.9$ & $71.4$ \\
\textbf{LaplacianShot} \cite{ziko2020laplacian} & \textbf{Standard} & $\mathbf{48.5}$ & $\mathbf{68.0}$& $\mathbf{72.8}$ \\
\hline & &  NCT \\
Method & Training Method & 5-way 1-shot & 5-way 5-shot & 5-way 10-shot \\
\hline MAML \cite{finn2017modelagnostic}  & Episodic & $57.2$ & $65.5$ & $69.2$ \\
ProtoNet \cite{snell2017prototypical} & Episodic & $62.6$ & $80.9$ & $84.9$ \\
DeepEMD \cite{zhang2020deepemd} & Episodic & $68.5$ & $84.0$ & $86.0$ \\
SimpleShot \cite{wang2019simpleshot} & Standard & $71.2$ & $85.5$ & $88.2$ \\
\textbf{LaplacianShot} \cite{ziko2020laplacian} & \textbf{Standard} & $\mathbf{71.8}$ & $\mathbf{86.9}$ & $\mathbf{89.5}$ \\
\hline &  \multicolumn{3}{c}{ LC25000 } \\
Method & Training Method & 5-way 1-shot & 5way 5-shot & 5-way 10-shot \\
\hline MAML \cite{finn2017modelagnostic}  &Episodic& $58.5$ & $69.2$ & $74.8$ \\
ProtoNet \cite{snell2017prototypical}  & Episodic & $67.2$ & $84.8$ & $86.2$ \\
DeepEMD \cite{zhang2020deepemd}  & Episodic & $73.8$ & $85.3$ & $86.4$ \\
SimpleShot \cite{wang2019simpleshot}  & Standard & $66.4$ & $83.6$ & $87.2$ \\
\textbf{LaplacianShot} \cite{ziko2020laplacian}  & \textbf{Standard} & $\mathbf{67.5}$ & $\mathbf{84.2}$ & $\mathbf{87.9}$ \\
\hline\
\end{tabular}
\end{center}
\end{table*}

\begin{figure}[htbp]
\includegraphics[height=6cm,width=8cm]{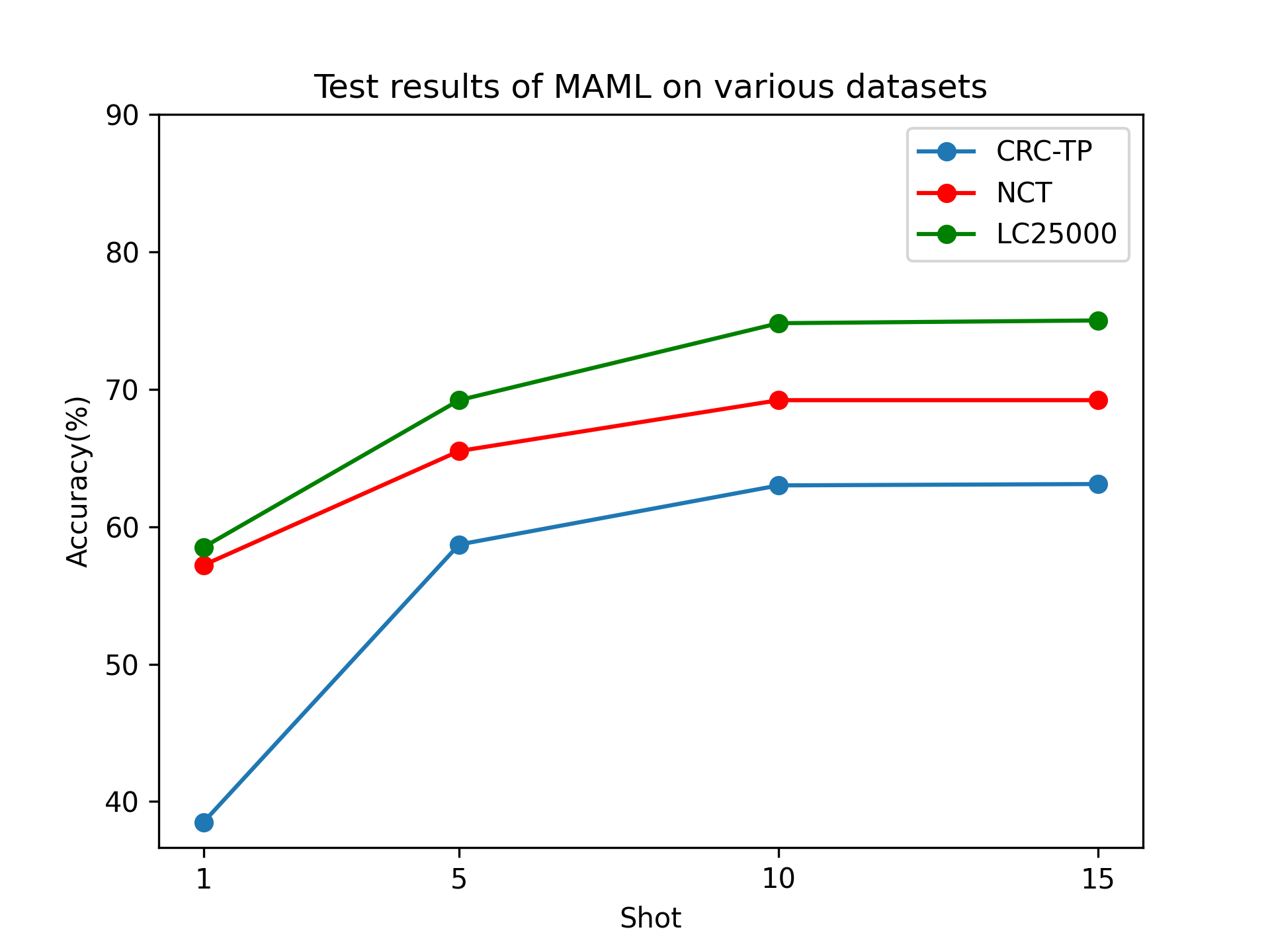}
\caption{Accuracy plot of MAML on different datasets}
\label{fig4}
\end{figure}
\begin{figure}[htbp]
\includegraphics[height=6cm,width=8cm]{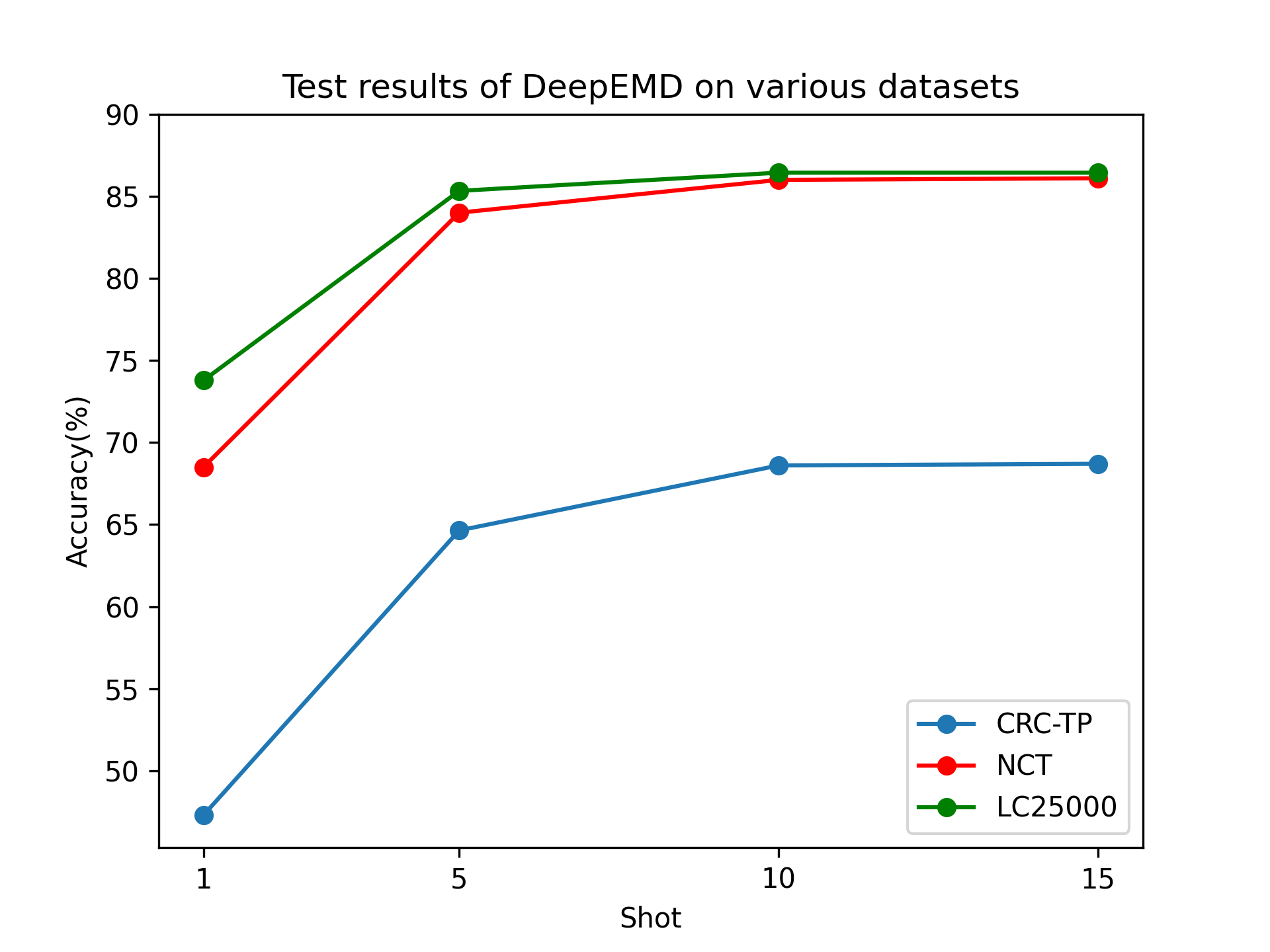}
\caption{Accuracy plot of DeepEMD on different datasets}
\label{fig5}
\end{figure}
\begin{figure}[htbp]
\includegraphics[height=6cm,width=8cm]{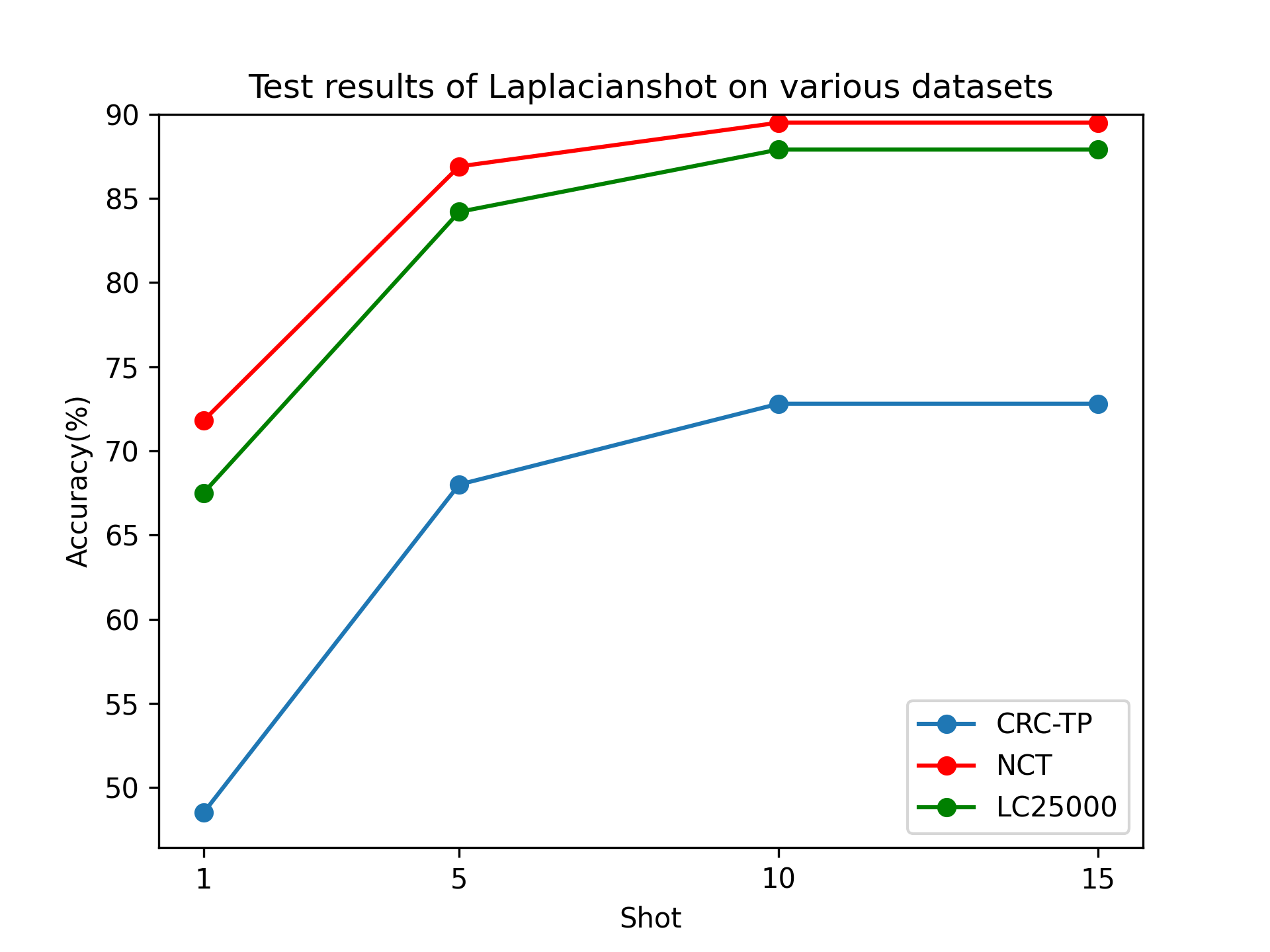}
\caption{Accuracy plot of Laplacianshot on different datasets}
\label{fig6}
\end{figure}

In episodic training, the methods are trained for 120 epochs. In each epoch, 600 episodes are randomly selected from the training set. The models are trained on 5-way 1-shot, 5-way 5-shot and 5-way 10-shot scenarios separately. In all these cases, the number of query images are set to 15. The initial learning rate is set to 1e-3. $\gamma$ is set to 0.1. The batch size is set to 1 episode. During 5-way 1-shot, 5-way 5-shot and 5-way 5-shot training, the number of images in a batch are (5$\times$1$+$5$\times$15) 80, (5$\times$5$+$5$\times$15) 100 and (5$\times$10$+$5$\times$15) 125 respectively. Few-shot methods that follow standard CNN training have been trained for 150 epochs. The batch size is set to 512. The initial learning rate, weight decay are initialized as 0.05 and 5e-4. Resnet18 is used as a backbone network in both episodic and training procedure.

All the trained models follow meta-testing. In this procedure, 5000 episodes are randomly sampled from CRC-TP, NCT and LC25000 dataset. In all K-shot testing scenarios, the number of query images are fixed as 15 in each episode. The results on 1-shot, 5-shot, 10-shot on different datasets are reported in the Table 1. The results we achieved by employing few-shot methods on the histology dataset align to a great extent with the outcomes obtained from the datasets of natural images, but there were a few surprises as well. As anticipated, few-shot techniques trained using standard fine-tuning such as SimpleShot and LaplacianShot yielded better results compared to approaches that adhered to episodic training such as MAML, ProtoNet and DeepEMD. The possible explanation for standard training procedures outperforming episodic training procedures is due to training of these methods on the substantial size of the dataset proposed by Komura et al., which includes cancer image patches from 32 diverse classes or organs. For histopathology images, the results obtained from ProtoNet are comparable to those achieved with DeepEMD. Considering that for natural images, DeepEMD exhibits significantly superior performance compared to ProtoNet. MAML, a few-shot learning-to-learn paradigm designed to learn and adapt quickly, does not appear to be well-suited for histopathology images.  As the number of shots are increased in all few-shot methods, there is an observed tendency for the results to reach a saturation point after 10 shots, as observed from the above plots. In 10-shot test scenarios, the accuracies of all methods, except MAML, across all datasets, excluding CRC-TP, fell within the range of 85\% to 90\%. In the context of complete training, \cite{javed2020cellular} reported an accuracy of 84.1\% on the CRC-TP dataset. Likewise, \cite{kather2019predicting} achieved an accuracy of 94.3\% on the NCT dataset, and \cite{sarwinda2020analysis} reported an accuracy of 98.5\% on the LC25000 dataset. The highest accuracy attained through the 10-shot method is comparable to that achieved through complete training.

\section{\uppercase{Conclusion}}

In conclusion, this research has explored the application of few-shot classification in the domain of histopathology images. Despite the growing prominence of few-shot learning in image classification, its application in histopathology has remained relatively uncharted. This study has specifically addressed the challenges arising from the scarcity of labeled data in medical imaging. Through the utilization of certain state-of-the-art few-shot learning methods, we evaluated their performance across various scenarios within the histology data domain. The consideration of four histopathology datasets has contributed significantly to our broader understanding of few-shot histopathology image classification. The evaluation process, scrutinizing 5-way 1-shot, 5-way 5-shot, and 5-way 10-shot scenarios, employing state-of-the-art classification techniques, has yielded remarkable results. The selected methods demonstrated exceptional capabilities, achieving accuracies close to 90\% in the 5-way 10-shot scenarios. The top accuracy obtained through the 10-shot method closely aligns with the accuracy achieved through full training, emphasising the effectiveness of Few-Shot Learning particularly when ample pre-training data is not accessible. These insights significantly contribute to the ongoing discourse surrounding the optimization of classification techniques for medical imaging, particularly in scenarios where labeled data is limited.

Future work in the application of few-shot classification in histopathology images could focus on multi-modal image classification, transfer learning, and pre-training, as well as the development of explainability and interpretability methods. Few-shot learning in multi-modal medical image classification involves training models to accurately classify medical images with limited examples per pathology or disease, leveraging information from various modalities like visual (X-rays or MRIs), textual (medical reports or descriptions), and potentially sensor data. In medical scenarios where obtaining a large labeled dataset for every pathology is challenging, this approach becomes crucial. By combining information from different modalities and addressing the challenges posed by limited examples, the model aims to make accurate predictions in medical image classification tasks, particularly where traditional deep learning models may struggle due to the scarcity of labeled medical data. Additionally, researchers can explore innovative data augmentation techniques and active learning strategies. Improving robustness to image variability, fostering collaboration with domain experts, establishing benchmark datasets, and addressing real-time implementation challenges are also crucial for advancing the field. Overall, these directions aim to enhance diagnostic accuracy and efficiency in histopathology through the effective use of limited labeled data and the integration of machine learning into clinical workflows.

\bibliographystyle{apalike}
{\small
\bibliography{example}}

\begin{thebibliography}{}

\bibitem[Borkowski et~al., 2019]{borkowski2019lung}
Borkowski, A.~A., Bui, M.~M., Thomas, L.~B., Wilson, C.~P., DeLand, L.~A., and Mastorides, S.~M. (2019).
\newblock Lung and colon cancer histopathological image dataset (lc25000).

\bibitem[Chen et~al., 2021]{chen2021momentum}
Chen, X., Yao, L., Zhou, T., Dong, J., and Zhang, Y. (2021).
\newblock Momentum contrastive learning for few-shot covid-19 diagnosis from chest ct images.
\newblock {\em Pattern Recognition}, 113:107826.

\bibitem[Finn et~al., 2017]{finn2017modelagnostic}
Finn, C., Abbeel, P., and Levine, S. (2017).
\newblock Model-agnostic meta-learning for fast adaptation of deep networks.
\newblock In {\em International Conference on Machine Learning (ICML)}.

\bibitem[Javed et~al., 2020a]{javed2020cellular}
Javed, S., Mahmood, A., Fraz, M.~M., Koohbanani, N.~A., Benes, K., Tsang, Y.-W., Hewitt, K., Epstein, D., Snead, D., and Rajpoot, N. (2020a).
\newblock Cellular community detection for tissue phenotyping in colorectal cancer histology images.
\newblock {\em Medical Image Analysis}, 63:101696.

\bibitem[Javed et~al., 2020b]{9204851}
Javed, S., Mahmood, A., Werghi, N., Benes, K., and Rajpoot, N. (2020b).
\newblock Multiplex cellular communities in multi-gigapixel colorectal cancer histology images for tissue phenotyping.
\newblock {\em IEEE Transactions on Image Processing}, 29:9204--9219.

\bibitem[Kather et~al., 2019]{kather2019predicting}
Kather, J.~N., Krisam, J., Charoentong, P., Luedde, T., Herpel, E., Weis, C.-A., Gaiser, T., Marx, A., Valous, N.~A., Ferber, D., Jansen, L., Reyes-Aldasoro, C.~C., Z{\"o}rnig, I., J{\"a}ger, D., Brenner, H., Chang-Claude, J., Hoffmeister, M., and Halama, N. (2019).
\newblock Predicting survival from colorectal cancer histology slides using deep learning: A retrospective multicenter study.
\newblock {\em PLoS Med}, 16(1):e1002730.

\bibitem[Kim et~al., 2021]{kim2021paip}
Kim, Y.~J., Jang, H., Lee, K., Park, S., Min, S.-G., Hong, C., Park, J.~H., Lee, K., Kim, J., Hong, W., et~al. (2021).
\newblock Paip 2019: Liver cancer segmentation challenge.
\newblock {\em Medical Image Analysis}, 67:101854.

\bibitem[Komura and Ishikawa, 2021]{komura2021histology}
Komura, D. and Ishikawa, S. (2021).
\newblock Histology images from uniform tumor regions in tcga whole slide images.
\newblock {\em Cell Reports}, 38(9):110424.

\bibitem[Mahajan et~al., 2020]{mahajan2020metadermdiagnosis}
Mahajan, K., Sharma, M., and Vig, L. (2020).
\newblock Meta-dermdiagnosis: Few-shot skin disease identification using meta-learning.
\newblock In {\em Proceedings of the IEEE/CVF Conference on Computer Vision and Pattern Recognition Workshops}, pages 730--731.

\bibitem[Medela et~al., 2019]{medela2019fewshot}
Medela, A., Picon, A., Saratxaga, C.~L., Belar, O., Cabezón, V., Cicchi, R., Bilbao, R., and Glover, B. (2019).
\newblock Few-shot learning in histopathological images: reducing the need of labeled data on biological datasets.
\newblock In {\em 2019 IEEE 16th International Symposium on Biomedical Imaging (ISBI 2019)}, pages 1860--1864. IEEE.

\bibitem[Nichol and Schulman, 2018]{nichol2018reptile}
Nichol, A. and Schulman, J. (2018).
\newblock Reptile: A scalable metalearning algorithm.
\newblock {\em arXiv preprint arXiv:1803.02999}.

\bibitem[Ouyang et~al., 2020]{ouyang2020selfsupervision}
Ouyang, C., Biffi, C., Chen, C., Kart, T., Qiu, H., and Rueckert, D. (2020).
\newblock Self-supervision with superpixels: Training few-shot medical image segmentation without annotation.
\newblock In {\em European Conference on Computer Vision}, pages 762--780. Springer.

\bibitem[Sarwinda et~al., 2020]{sarwinda2020analysis}
Sarwinda, D., Bustamam, A., Paradisa, R.~H., Argyadiva, T., and Mangunwardoyo, W. (2020).
\newblock Analysis of deep feature extraction for colorectal cancer detection.
\newblock In {\em 2020 4th International Conference on Informatics and Computational Sciences (ICICoS)}, pages 1--5. IEEE.

\bibitem[Shakeri et~al., 2022]{shakeri2022fhist}
Shakeri, F., Boudiaf, M., Mohammadi, S., Sheth, I., Havaei, M., Ben~Ayed, I., and Kahou, S.~E. (2022).
\newblock Fhist: A benchmark for few-shot classification of histological images.
\newblock {\em arXiv}.

\bibitem[Snell et~al., 2017]{snell2017prototypical}
Snell, J., Swersky, K., and Zemel, R. (2017).
\newblock Prototypical networks for few-shot learning.
\newblock In {\em Advances in Neural Information Processing Systems (NeurIPS)}.

\bibitem[Spanhol et~al., 2016]{7312934}
Spanhol, F.~A., Oliveira, L.~S., Petitjean, C., and Heutte, L. (2016).
\newblock A dataset for breast cancer histopathological image classification.
\newblock {\em IEEE Transactions on Biomedical Engineering}, 63(7):1455--1462.

\bibitem[tcga, 2005]{tcga}
tcga (2005).
\newblock The cancer genome atlas.
\newblock https://www.cancer.gov/ccg/research/genome-sequencing/tcga.
\newblock Accessed on: 2023-11-30.

\bibitem[Teng et~al., 2021]{teng2021fewshot}
Teng, H., Zhang, W., Wei, J., Lv, L., Tang, L., Fu, C.-C., Cai, Y., Qin, G., Ye, M., and Fang, Qu, e.~a. (2021).
\newblock Few-shot learning on the diagnosis of lymphatic metastasis of lung carcinoma.
\newblock {\em Research Square}.

\bibitem[Wang et~al., 2019]{wang2019simpleshot}
Wang, Y., Chao, W.-L., Weinberger, K.~Q., and van~der Maaten, L. (2019).
\newblock Simpleshot: Revisiting nearest-neighbor classification for few-shot learning.
\newblock {\em arXiv preprint arXiv:1911.04623}.

\bibitem[Yang et~al., 2022]{yang2022towards}
Yang, J., Chen, H., Yan, J., Chen, X., and Yao, J. (2022).
\newblock Towards better understanding and better generalization of few-shot classification in histology images with contrastive learning.
\newblock {\em arXiv preprint arXiv:2202.09059}.

\bibitem[Yu et~al., 2021]{yu2021locationsensitive}
Yu, Q., Dang, K., Tajbakhsh, N., Terzopoulos, D., and Ding, X. (2021).
\newblock A location-sensitive local prototype network for few-shot medical image segmentation.
\newblock In {\em 2021 IEEE 18th International Symposium on Biomedical Imaging (ISBI)}, pages 262--266. IEEE.

\bibitem[Zhang et~al., 2020]{zhang2020deepemd}
Zhang, C., Cai, Y., Lin, G., and Shen, C. (2020).
\newblock Deepemd: Few-shot image classification with differentiable earth mover's distance and structured classifiers.
\newblock In {\em Conference on Computer Vision and Pattern Recognition (CVPR)}.

\bibitem[Ziko et~al., 2020]{ziko2020laplacian}
Ziko, I.~M., Dolz, J., Granger, E., and Ben~Ayed, I. (2020).
\newblock Laplacian regularized few-shot learning.
\newblock In {\em International Conference on Machine Learning (ICML)}.

\end{thebibliography}

\end{document}